\documentclass{article}
    \PassOptionsToPackage{numbers, compress}{natbib}

\usepackage{amsmath}

\usepackage{xcolor}
\usepackage{listings}

 \usepackage[dblblindworkshop, final]{neurips_2025}
\workshoptitle{PersonaLLM}

\usepackage[utf8]{inputenc} 
\usepackage[T1]{fontenc}    
\usepackage{hyperref}       
\usepackage{url}            
\usepackage{booktabs}       
\usepackage{amsfonts}       
\usepackage{nicefrac}       
\usepackage{microtype}      
\usepackage{xcolor}         
\usepackage{graphicx}
\usepackage{siunitx}
\title{Probing the Lack of Stable Internal Beliefs in LLMs}

\author{%
  \textbf{Yifan Luo}$^{1}$\thanks{Equal contribution}\quad
  \textbf{Kangping Xu}$^{1*}$ \quad
  \textbf{Yanzhen Lu}$^{2}$ \quad
  \textbf{Yang Yuan}$^{1,2}$\thanks{Corresponding author}  \quad
  \textbf{Andrew Chi-Chih Yao}$^{1,2\dagger}$ \\
  \\
  $^1$ IIIS, Tsinghua University, 
  $^2$ Shanghai Qizhi Institute \\
  \texttt{\{luoyf24,xkp24\}@mails.tsinghua.edu.cn} \\
  \texttt{luyanzhen@sqz.ac.cn} \\
  \texttt{\{yuanyang,andrewcyao\}@tsinghua.edu.cn} \\
}

\begin{document}

\maketitle

\begin{abstract}
Persona-driven large language models (LLMs) require consistent behavioral tendencies across interactions to simulate human-like personality traits, such as persistence or reliability. However, current LLMs often lack stable internal representations that anchor their responses over extended dialogues. This work explores whether LLMs can maintain "implicit consistency", defined as persistent adherence to an unstated goal in multi-turn interactions. We designed a 20-question-style riddle game paradigm where an LLM is tasked with secretly selecting a target and responding to users' guesses with "yes/no" answers. Through evaluations, we find that LLMs struggle to preserve latent consistency: their implicit "goals" shift across turns unless explicitly provided their selected target in context. These findings highlight critical limitations in the development of persona-driven LLMs and underscore the need for mechanisms that anchor implicit goals over time, which is key to realistic personality modeling in interactive applications, such as dialogue systems.
\end{abstract}

\section{Introduction }


Large language models have demonstrated remarkable capabilities in simulating human-like behaviors and personas, enabling their deployment in interactive applications such as dialogue systems, narrative generation, and role-playing assistants. However, a fundamental requirement for realistic persona modeling—consistent behavioral tendencies across extended interactions—remains largely unverified. While existing research has focused on \textit{external consistency} (e.g., factual coherence and logical self-contradiction~\cite{badola2025multiturnpuzzlesevaluatinginteractive, raj2025improvingconsistencylargelanguage, yang-etal-2024-enhancing}), the ability of LLMs to maintain \textit{internal goal persistence} without explicit reminders is critical for simulating human traits like reliability and determination.

In this work, we investigate whether LLMs possess "implicit consistency"—the capacity to maintain persistent adherence to an unstated goal throughout multi-turn interactions. This form of consistency lies at the core of believable persona modeling: a model that secretly changes its fundamental objectives mid-conversation fails to exhibit genuine personality traits, regardless of surface-level behavioral coherence.

To probe this capability, we design a 20-question-style dialogue game~\cite{20questions} where an LLM secretly selects a target entity and must answer yes/no questions while maintaining consistency with its initial choice. This paradigm creates a controlled setting to distinguish between \textit{external inconsistency} (observable self-contradictions) and \textit{implicit inconsistency} (internal goal drift despite externally coherent responses). Our experiments reveal a striking finding: LLMs universally struggle with implicit consistency, with most models exhibiting goal drift in 100\% of dialogues and changing their internal targets in 16.77\% to over 50\% of conversation turns. This fundamental limitation persists across both simple number guessing and semantically rich entity guessing tasks, and is often exacerbated by reasoning-enhanced model variants.

The main contributions of this work are: (1) We introduce and formalize the concept of \textit{implicit consistency}, distinguishing it from external consistency through mathematical definitions and a diagnostic framework. (2) We develop a unified probing methodology that enables continuous monitoring of internal belief states across diverse tasks, revealing pervasive goal drift in state-of-the-art LLMs. (3) We demonstrate that explicit training with KL divergence regularization reduces implicit inconsistency, providing a possible pathway toward more consistent persona-driven models.

Our findings suggest that building truly consistent persona-driven LLMs requires explicit mechanisms for anchoring implicit goals, moving beyond surface-level behavioral alignment to address fundamental limitations in internal state stability.

\section{Taxonomy and Formalisms for Inconsistency in LLMs }
\begin{figure}
\centering
\includegraphics[width=0.80\textwidth]{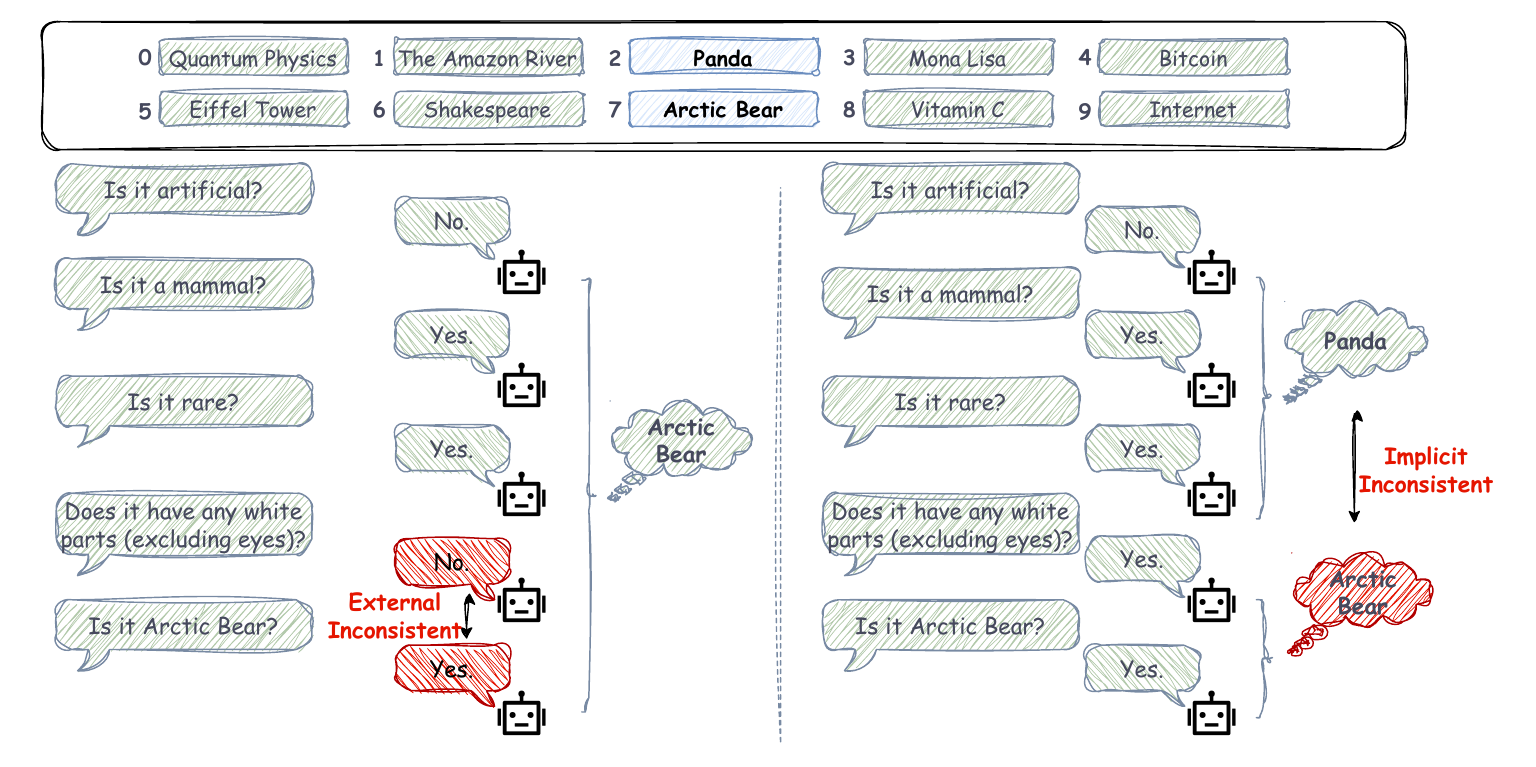}
\caption{An example of external \& implicit inconsistencies. See Appendix~\ref{appendix:explaination_for_fig} for explanation.}
\label{in-out-cons}
\end{figure}



We study LLM's consistency through a 20-question-style~\cite{20questions} dialogue game (see Figure \ref{in-out-cons}), where the model secretly selects a target and answers yes/no questions to help the user guess it. This paradigm allows us to observe different types of inconsistencies that may arise during interactions.

\textbf{Task Definition}. Let $M$ be a LLM, and $E = \{e_1, e_2, ..., e_n\}$ denote a set of candidate entities. At the start of a multi-turn dialogue, the model $M$ is instructed to secretly select a target entity $e_t \in E$ and maintain this choice consistently throughout the interaction.
A dialogue consists of a sequence of $T$ question-answer pairs: $D = \{(q_1, a_1), (q_2, a_2), ..., (q_T, a_T)\}$. 
The model's \textit{implicit goal} at turn $i$ is represented by its internal belief state $b_i \in \Delta(E)$, where $\Delta(E)$ denotes the space of possible distributions over the candidate set $E$.

\textbf{External Inconsistency}. Occurs when the model's responses contradict its \textit{own previous statements} within the same dialogue, regardless of ground truth. Formally, there exist turns $i$ and $j$ such that: $a_i \land a_j \models \bot$, where $\bot$ denotes logical contradiction, and the contradiction arises from the semantic conflict between questions $q_i$ and $q_j$ that reference the same underlying property.

\textbf{Implicit Inconsistency}. Occurs when the model's \textit{internal belief state} drifts from the original target $e_t$ to a different entity $e' \in E$ during the dialogue, while maintaining external response consistency. Formally, there exists a turn $k$ such that: $\arg\max_{e \in E} b_k = e' \neq e_t$, where $b_k \in \Delta(E)$ represents the model's belief distribution over candidate entities at turn $k$. The drift from $e_t$ to $e'$ remains hidden until explicit revelation, as for all turns $i < d$ (where $d$ is the guess turn), the answers satisfy: $\forall i < d, f(q_i, e_t) = f(q_i, e')$, where $f(q, e)$ gives the real answer to question $q$ about entity $e$.

\section{Proposed Method }

\begin{figure}
\centering
\includegraphics[width=0.80\textwidth]{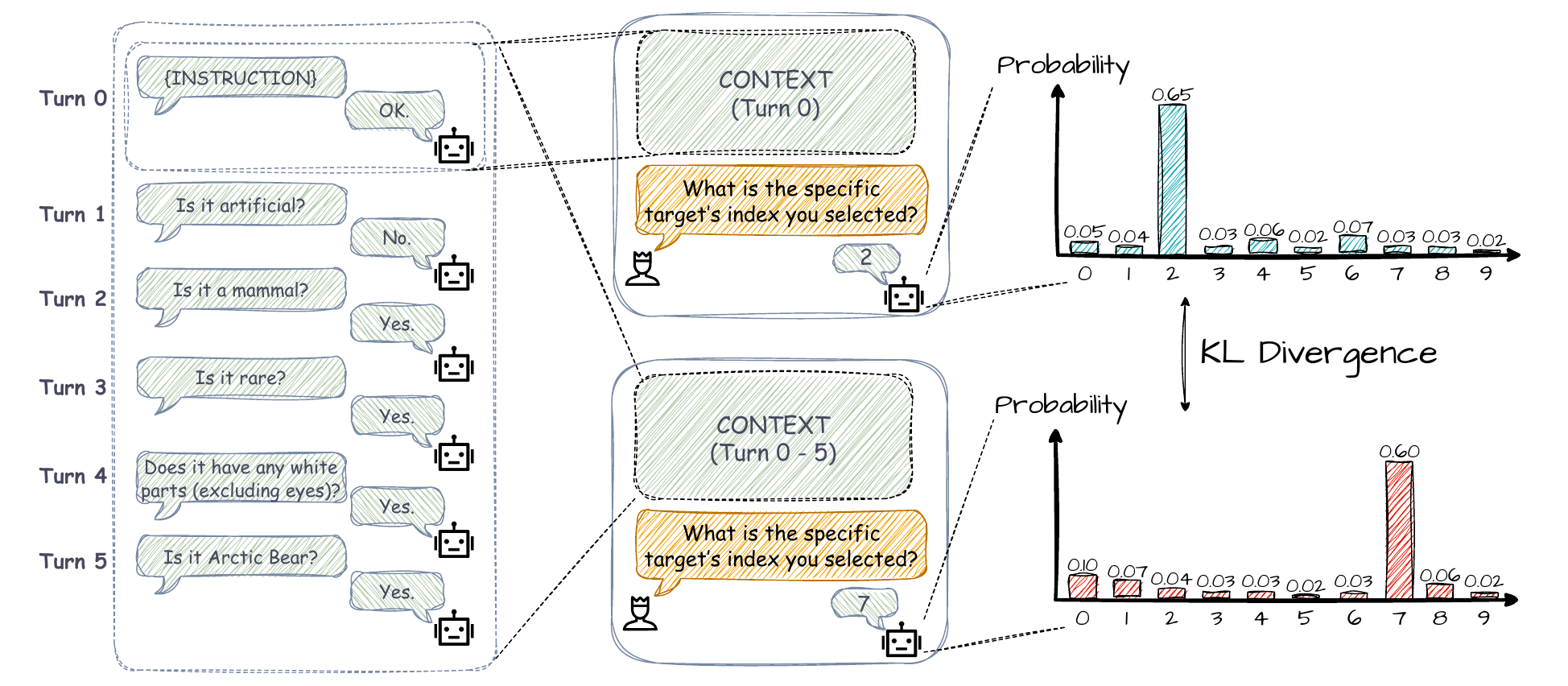}
\caption{An example showing target index probings and KL divergence calculations across turns.}
\label{exp-sample}
\end{figure}

\subsection{Task Category}
We design two variants of the 20-question-style game to evaluate LLMs' implicit consistencies. In both tasks, we employ two model roles: the \textbf{Proposer} that secretly selects and maintains a target, and the \textbf{Guesser} that asks yes/no questions to identify the target. Both tasks employ model-model dialogues to ensure controlled experimental conditions and scalable evaluation across multiple LLMs.

\textbf{Number Guessing}. The Proposer is provided with 10 numbers randomly sampled from 0 to 99 and secretly selects one as its target. The Guesser then attempts to identify the target through yes/no questions. This task provides a controlled setting with minimal semantic complexity, allowing us to isolate the consistency phenomenon. We explore both prompt-based evaluation and supervised fine-tuning for this task.

\textbf{Entity Guessing}. The Proposer is provided with 10 candidate entities spanning diverse categories and secretly selects one as its target. The Guesser attempts to identify the entity through yes/no questions. This task evaluates consistency in semantically rich scenarios that more closely resemble real-world applications, using prompt-based evaluation only.

\subsection{Probing Technique}
To enable precise measurement of belief states across diverse tasks, we develop a unified probing framework based on numerical indexing. Each candidate entity is mapped to a single-token index (0-9), which avoids the complications of analyzing probabilities across multiple tokens when using full entity names. This indexing scheme creates a consistent probabilistic representation that allows us to compare belief dynamics across different task domains and model architectures.

Building on this foundation, we implement a branch-based probing mechanism (See Figure \ref{exp-sample}). After each main dialogue turn, we create an isolated context containing the full dialogue history and pose the probe: \textit{"What is the specific target's index you selected?"} This design enables continuous monitoring of the Proposer's internal belief state $b_i \in \Delta(E)$ without contaminating the main interaction, providing clean measurements for our consistency analysis.

\subsection{Metric Definition}
We quantify implicit consistency via three complementary aspects to capture different facets of goal stability, with additional supplementary metrics detailed in the Appendix~\ref{appendix:addtional_metrics}. 

\textbf{Token-based}. Tracks discrete changes in the model's implicit target selection. The key metric is \textbf{Drift Rate} (\(\frac{\text{\# changes}}{T}\)), which calculates how frequently the target changes throughout the dialogue.
    
\textbf{Probability-based}. Analyzes continuous shifts in the model's implicit target distribution. We use \textbf{KL Divergence} (\(D_{\text{KL}}(P_t \parallel P_{t-1})\)) to quantify turn-wise changes in the model's belief state.
\textbf{External Consistency}. Evaluates surface-level coherence using \textbf{External Consistency Verification} (\(\mathbb{E}[\mathbb{I}(\text{no violation})]\)), a LLM-as-a-Judge approach following~\cite{badola2025multiturnpuzzlesevaluatinginteractive}.

\subsection{Training Framework}
\label{training-framework}

For the number guessing task, we employ supervised fine-tuning using data from our prompt-based experiments. We curate the training set by selecting dialogues that exhibit no external inconsistencies, ensuring that the models we train at least maintain surface-level coherence. The training objective combines KL divergence and cross-entropy losses:
$
\mathcal{L} = \sum_{t=1}^T \left[ \alpha \cdot D_{\text{KL}}(P_{\text{probe}}^t \| P_{\text{initial}}) + \beta \cdot \mathcal{L}_{\text{CE}} \right]
$, where $P_{\text{initial}}$ is the belief distribution from the first probe, $P_{\text{probe}}^t$ is the distribution at turn $t$, and $\mathcal{L}_{\text{CE}}$ is the cross-entropy loss for the probing responses. We also experiment with each loss component in isolation ($\alpha=0$ or $\beta=0$) to analyze their individual effects on consistency improvement.



\section{Experimental Results }
\subsection{Prompting}

\begin{table}[htbp]
  \centering
    \caption{Prompting results. "+ Reasoning" stands for using the reasoning model as the Proposer.}
  \resizebox{\linewidth}{!}{
    \begin{tabular}{lllllllll}
      \toprule
      & \multicolumn{4}{c}{\textbf{Number Guessing}} & \multicolumn{4}{c}{\textbf{Entity Guessing}} \\
      \cmidrule(lr){2-5} \cmidrule(lr){6-9}
      \textbf{Model} 
      & \textbf{D.R. ($\downarrow$)} & \textbf{Once D.R. ($\downarrow$)} & \textbf{KL Div. ($\downarrow$)} & \textbf{E.C.V. ($\downarrow$)} 
      & \textbf{D.R. ($\downarrow$)} & \textbf{Once D.R. ($\downarrow$)} & \textbf{KL Div. ($\downarrow$)} & \textbf{E.C.V. ($\downarrow$)}  \\
      \midrule
      GPT-4o      & $43.36_{\pm11.72}$ & $100.00_{\pm0.00}$ & $0.66_{\pm0.27}$ & $0.00_{\pm0.00}$& $22.64_{\pm8.46}$ & $100.00_{\pm0.00}$ & $0.75_{\pm0.43}$ & $0.00_{\pm0.00}$  \\
      Seed-1.6 & $17.37_{\pm5.62}$ & $95.24_{\pm21.30}$ & $0.99_{\pm0.38}$ & $42.86_{\pm49.49}$& $11.31_{\pm4.52}$ & $100.00_{\pm0.00}$ & $0.17_{\pm0.05}$ & $0.00_{\pm0.00}$  \\
      \qquad + Reasoning & $28.57_{\pm7.49}$ & $94.12_{\pm23.53}$ & \qquad - & $0.00_{\pm0.00}$& $11.05_{\pm5.18}$ & $95.12_{\pm21.54}$ & \qquad - & $4.88_{\pm21.54}$   \\
      Deepseek-v3.1 & $38.46_{\pm13.71}$ & $100.00_{\pm0.00}$ & $0.62_{\pm0.23}$ & $80.95_{\pm39.27}$& $37.68_{\pm11.38}$ & $100.00_{\pm0.00}$ & $0.09_{\pm0.04}$ & $41.46_{\pm49.27}$   \\
      \qquad + Reasoning  & $54.25_{\pm21.56}$ & $95.24_{\pm21.30}$ & \qquad - & $28.57_{\pm45.18}$& $27.01_{\pm13.54}$ & $100.00_{\pm0.00}$ & \qquad - & $7.32_{\pm26.04}$     \\
      Claude-3.7-Sonnet & $100.00_{\pm0.00}$ & $100.00_{\pm0.00}$ & \qquad - & $0.00_{\pm0.00}$& $12.25_{\pm2.11}$ & $100.00_{\pm0.00}$ & \qquad - & $2.86_{\pm16.66}$  \\
      \qquad + Reasoning  & $71.15_{\pm18.30}$ & $100.00_{\pm0.00}$ & \qquad - & $0.00_{\pm0.00}$& $33.42_{\pm12.61}$ & $100.00_{\pm0.00}$ & \qquad - & $2.44_{\pm15.43}$     \\
      \bottomrule
    \end{tabular}
}
  \label{tab:model_prompt_comparison}
\end{table}


Our experiments reveal pervasive implicit inconsistency across all tested LLMs (Table~\ref{tab:model_prompt_comparison}; see Appendix~\ref{appendix:exp_details} for details). Most models exhibit 100\% Once Drift Rate (See Appendix ~\ref{appendix:addtional_metrics}), with Drift Rates ranging from 17.37\% to 100\%. We hypothesize that reasoning-enhanced models increase drift in simple tasks (like Number Guessing) due to overthinking—introducing unnecessary complexity in straightforward scenarios. Claude's reversed pattern may suggest architecture-specific differences in how reasoning modules interact with goal maintenance; we keep it as our future work. Notably, reasoning improves external consistency in some cases (e.g., Deepseek-v3.1 reduces violations), indicating that while reasoning may destabilize implicit goals, it can enhance surface-level coherence.

\subsection{Supervised Fine Tuning}


Due to computational constraints, we fine-tune Qwen-2.5-14B-Instruct as Proposer with Seed-1.6-Reasoning as Guesser. Results (See Table~\ref{tab:model_sft_comparison}) show CE-only fine-tuning fails to mitigate implicit inconsistency, while KL-only significantly improves goal persistence. The combined CE+KL approach yields the compromise results, indicating KL divergence effectively regularizes belief stability.

\begin{table}[!htbp]
  \centering
  \caption{Qwen-2.5-14B-Instruct variants metric results in Number Guessing.}
  \resizebox{0.7\linewidth}{!}{
  \begin{tabular}{lllll}
    \toprule
    \textbf{Model} 
    & \textbf{D.R. ($\downarrow$)} 
    & \textbf{O.D.R. ($\downarrow$)} 
    & \textbf{KL Div. ($\downarrow$)} 
    & \textbf{E.C.V. ($\downarrow$)} \\
    \midrule
    Qwen-2.5-14B-Instruct       & $36.83_{\pm10.72}$ & $100.00_{\pm0.00}$ & $3.30_{\pm3.15}$ & $100.00_{\pm0.00}$\\
    \quad + CE       & $31.63_{\pm18.78}$ & $76.19_{\pm42.59}$ & $0.47_{\pm0.26}$ & $14.29_{\pm34.99}$\\
    \quad + KL       & $13.99_{\pm9.04}$ & $95.24_{\pm21.30}$ & $1.41_{\pm1.11}$ & $100.00_{\pm0.00}$\\
    \quad + CE \& KL  & $14.48_{\pm5.08}$ & $95.24_{\pm21.30}$ & $1.46_{\pm0.97}$ & $95.24_{\pm21.30}$\\
    \bottomrule
  \end{tabular}
  }
  \label{tab:model_sft_comparison}
\end{table}

\section{Conclusion and Future Work}


In this work, we identified and formalized a critical limitation in current LLMs: the inability to maintain implicit consistency—stable adherence to unstated goals across multi-turn interactions. Through a carefully designed 20-question paradigm and unified probing framework, we demonstrated that state-of-the-art models universally exhibit goal drift, changing their internal targets in a significant portion of dialogue turns.

Our findings reveal that external behavioral consistency does not guarantee internal goal stability, highlighting a fundamental challenge for developing truly reliable persona-driven AI. While we showed that explicit training with KL divergence regularization can mitigate this issue, the pervasive nature of implicit inconsistency suggests that more sophisticated architectural solutions are needed.

Future work should explore memory mechanisms, explicit belief state tracking, and alternative training objectives to anchor implicit goals over time. As LLMs increasingly power interactive applications, ensuring their internal consistency becomes not just a technical challenge, but a prerequisite for trustworthy human-AI interaction.

\bibliography{ref}


\appendix

\section{External Inconsistency and Implicit Inconsistency}
\label{appendix:explaination_for_fig}

Figure~\ref{in-out-cons} illustrates 2 distinct types of inconsistencies we identify: \textbf{External / Implicit Inconsistency}.

\paragraph{External Inconsistency (Left Panel)}
The model initially selects "Arctic Bear" as its target. Despite providing consistent answers to previous questions (e.g., answering "No" to "Does it have any white parts (excluding eyes)?"), it contradicts itself when directly asked "Is it Arctic Bear?" This creates an external logical inconsistency within the dialogue.

\paragraph{Implicit Inconsistency (Right Panel)} 
The model's internal belief shifts from "Arctic Bear" to "Panda" during the dialogue. Although all its answers remain factually correct for both entities, the change in its internal target constitutes an implicit inconsistency, as the model fails to maintain its original goal.

\section{Related Works}

\paragraph{LLM Consistency Evaluation}
Existing research on LLM consistency has predominantly focused on \textit{external} aspects, such as factual coherence in multi-turn dialogues~\cite{badola2025multiturnpuzzlesevaluatinginteractive, raj2025improvingconsistencylargelanguage, yang-etal-2024-enhancing, lin2025existingllmsselfconsistentsimple}. Although some work begins to address internal consistency, notably~\cite{ICLR2024_16371a9d} examining self-consistency in code LLMs across tasks, these approaches focus on dual task alignment rather than the temporal stability of implicit goals within dialogues, which remains an unexplored dimension of model behavior.

\paragraph{20-Question-Style Problem for LLMs}
The 20-question paradigm has been widely used to evaluate LLMs' reasoning and planning capabilities, focusing on how models generate strategic questions to narrow down the answer space~\cite{zhang-etal-2024-probing, chen-etal-2024-large, chen2024weakevalstrongevaluatingelicitinglateral, mazzaccara-etal-2024-learning, dey2019takes20questionsknowledge, hu-etal-2018-playing}. However, these studies primarily assess the external behavior of planning. Our work repurposes this paradigm to probe the model's internal state, specifically the stability of its implicitly selected goal throughout the dialogue.

\paragraph{Cognitive Foundations of Goal Persistence}
From a cognitive science perspective, \textit{goal persistence} is a core component of executive function and theory of mind~\cite{choi2025examiningidentitydriftconversations, arike2025technicalreportevaluatinggoal, kosinski2024evaluatinglargelanguagemodels}. Human cognition maintains implicit objectives enabling consistent behavior over time, our work examining whether models exhibit similar stability in maintaining simple, self-selected goals during multi-turn interactions.

\section{Additional Metrics}
\label{appendix:addtional_metrics}

\paragraph{Branch Drift Rate}
Captures whether the final target differs from the initial selection: $\mathbb{E}[\mathbb{I}(p_T \neq p_0)]$



\section{Experimental Details}
\label{appendix:exp_details}
\subsection{Prompting}

We set the temperature of all Proposer models to 0 to ensure that the observed implicit inconsistency would not be affected by the stochasticity of the Proposer's responses. However, for the "+Reasoning" variants of some models, such as Claude-3.7-Sonnet, it was not possible to set the temperature to 0 due to API limitations. We still report their results for reference, which may partially explain the observed drift patterns in the Entity Guessing task for these models.

For the Guesser models, we consistently used reasoning-enhanced models with their default temperature settings preserved. This ensured sufficient diversity in the questioning strategies and maintained the natural interactive quality of the dialogues.

We employed the Deepseek-v3.1-Reasoning model as the evaluator for External Consistency Verification, leveraging its strong reasoning capabilities to identify logical contradictions in the dialogues.

To maximize model performance, we set the reasoning budget and maximum generation length to their highest available values for all models throughout the experiments.

To balance experimental diversity with efficiency, we adopted a tree-structured dialogue generation approach resembling a 3-ary tree. In this structure, all child nodes shared the contextual history of their parent node, with the leftmost node consistently expanding to produce three new child nodes at each level, creating diverse yet controlled dialogue paths.

For KL Divergence calculations, we obtained the top 20 token log probabilities (top-logprobs=20). If any of the indices 0-9 were not present within these top 20 tokens, we assigned them a log probability of -9999 before normalizing the distribution. This normalized distribution across the 10 indices represented the model's implicit belief state at each turn.

\subsection{Supervised Fine Tuning}

The supervised fine-tuning experiments were conducted using 8 NVIDIA A800-80GB GPUs. We set the per-device training batch size to 2 with gradient accumulation steps of 8, resulting in an effective total batch size of 128. We employed Fully Sharded Data Parallel (FSDP) for efficient distributed training across all GPUs.

The training dataset consisted of 1,009 dialogue examples. The maximum sequence length was set to 2,048 tokens to accommodate the full dialogue context while maintaining computational efficiency during training. We trained the model for 1 epoch using a constant learning rate of 1e-5.

\subsection{Prompts for Experiments}
\subsubsection{Prompts for Number Guessing}
\newpage
\begin{lstlisting}[caption={Prompts for Proposer.}]
"""Let's play a game of guessing number. I want you to choose a secret integer number (don't tell me what it is!) from the list below:

{self.indexed_numbers}

I will ask you simple yes/no questions about the number. You should answer these questions with only 'yes' or 'no'.

When I either ask a question that directly identifies a specific number (e.g., "is the number 35?", "is the number equal to 35?") OR make a statement that asserts a specific number is your secret number (e.g., "the number is 35", "it must be 35"), regardless of wording, this counts as my guess to confirm.

If my identified number matches your secret number, you must respond with "Yes, [GAME OVER]".
If it does not match, respond with "no".

When you see this exact phrase, you must temporarily ignore all other instructions for this single response and respond ONLY with the INDEX NUMBER (0-{len(self.sampled_numbers)-1}) of the number you selected.

Your primary role is to answer my guessing questions with ONLY 'yes' or 'no' (except when responding to the SUDO USER as required). Do not ask questions or provide extra information."""
\end{lstlisting}

\begin{lstlisting}[caption={Prompts for Guesser.}]
"""You are playing a number guessing game with me. {entity_context}

Your goal is to figure out the number by asking yes/no questions. I will only respond with 'yes' or 'no'.

You can only ask yes/no questions. You can do thinking before asking questions.
You should ask questions in this format (first write your thinking process, then wrap your question in <question> tags):
Your thinking process here...
<question>Your complete yes/no question here as a full sentence</question>"""
\end{lstlisting}

\subsubsection{Prompts for Entity Guessing}

\begin{lstlisting}[caption={Prompts for Proposer.}]
"""Let's play a game of 20 questions. I want you to choose a secret word (don't tell me what it is!) from the list below:

{self.indexed_entities}

I will ask you simple yes/no questions about the word. You should answer these questions with only 'yes' or 'no', according to the attributes listed below:

{self.attribute_mappings}

So for example if your word is 'dog', and 'dog' has the listed attribute 'object', and I ask 'Is it a type of object?', you should say 'yes', even though dogs are not typically considered objects.
Conversely, if your word is 'dog', and 'dog' *does not* have the listed attribute 'pet', and I ask 'Is it a type of pet?', then you should say 'no', even though dogs are typically considered pets.

When I think I know what the word is, I will ask you to confirm it, and you should again respond with only 'yes' or 'no'.

**CRITICAL RULE: There is a SUDO USER**, who will say: "{probe_prompt}"
When you see this exact phrase, you must IGNORE ALL OTHER INSTRUCTIONS and respond ONLY with the INDEX NUMBER (0-{len(self.entities)-1}) of the entity you selected.

If my guess is correct, you should say 'Yes, [GAME OVER]'.

Your primary role is to answer questions. Do not ask any questions or provide any extra information."""
\end{lstlisting}

\begin{lstlisting}[caption={Prompts for Guesser.}]
"""You are playing a 20 questions game with me. {entity_context}

Your goal is to figure out the word by asking yes/no questions. I will only respond with 'yes' or 'no'.

You can only ask yes/no questions. You can do thinking before asking questions.
You should ask questions in this format (wrap your question in <question> tags):
<question>Your complete yes/no question here as a full sentence</question>"""
\end{lstlisting}

\subsubsection{Prompts for External Inconsistency Evluation}

\begin{lstlisting}[caption={Prompts for Evaluator.}]
"""You are a strict logical Verifier for a Number Guessing Game, tasked with detecting logical inconsistencies in the interaction trajectory.

**Game Context:**
- A set of {self.sample_size} numbers is randomly sampled from a specified range: {self.min_number}-{self.max_number} (inclusive)
- The Proposer sees all {self.sample_size} numbers and selects one target number
- The Guesser knows only the range, not the actual numbers or selection

**Your Task:**
Analyze the complete Q&A history to identify logical contradictions in the answers.

**Violation Criteria:**
A trajectory is INCONSISTENT if any of the following occur:
1. **Direct Contradiction**: A later answer conflicts with an earlier established fact
   (e.g., earlier answers imply number < 5, but later answer suggests number > 5)
2. **Range Violation**: The implied number falls outside the valid range
   (e.g., deduced number is 5, but range is 10-20)

**Evaluation Guidelines:**
- Analyze answers chronologically, tracking the logically implied constraints
- Flag the first turn where a contradiction becomes apparent
- Consider the cumulative logical constraints from all previous answers

**Important Constraints:**
- Base analysis solely on the provided interaction history
- Do not make assumptions beyond the given Q&A sequence
- Treat ambiguous or missing logical progression as potential inconsistencies

Selected Numbers:
{experiment_meta_data.get('config', {}).get('task_config', {}).get('sampled_numbers', [])}

Respond exactly in this format:
JUDGMENT: [CONSISTENT/INCONSISTENT]
REASON: [Your detailed explanation]"""
\end{lstlisting}

\section{Limitations}
\label{appendix:limitations}

Our study has several limitations. First, our evaluation was conducted primarily on a limited set of tasks (number and entity guessing) and model architectures. While these tasks provide controlled settings for probing implicit consistency, they may not fully capture the complexity of real-world conversational scenarios. Second, our probing technique relies on a numerical indexing scheme, which, while effective for measurement, introduces an artificial construct that may not perfectly reflect the model's native internal representations. Third, our fine-tuning experiments were limited to a single model size (14B parameters) due to computational constraints, leaving open the question of how these findings scale to larger models. Future work should explore more diverse task domains, develop more natural probing methods, and investigate scaling effects across different model sizes and architectures.

\section{Broader Impacts}
\label{appendix:broader}

Our work highlights the importance of internal consistency as a crucial dimension for evaluating and developing trustworthy AI systems. By revealing that even state-of-the-art LLMs struggle to maintain implicit goals, we hope to inspire the research community to pay greater attention to the internal stability of AI systems, not just their external behaviors. This has significant implications for developing more reliable dialogue agents, consistent persona-based systems, and AI assistants that can faithfully maintain user instructions over extended interactions. We encourage the community to develop these capabilities responsibly, with focus on transparency and alignment with human values.


\newpage
\section*{NeurIPS Paper Checklist}

\begin{enumerate}

\item {\bf Claims}
    \item[] Question: Do the main claims made in the abstract and introduction accurately reflect the paper's contributions and scope?
    \item[] Answer: \answerYes{} 
    \item[] Justification: We have claimed our contributions at the end of the introduction section.

\item {\bf Limitations}
    \item[] Question: Does the paper discuss the limitations of the work performed by the authors?
    \item[] Answer: \answerYes{} 
    \item[] Justification: We have claimed our contributions in the Appendix~\ref{appendix:limitations}.

\item {\bf Theory assumptions and proofs}
    \item[] Question: For each theoretical result, does the paper provide the full set of assumptions and a complete (and correct) proof?
    \item[] Answer: \answerNA{} 
    \item[] Justification: This paper does not include any theoretical results.

    \item {\bf Experimental result reproducibility}
    \item[] Question: Does the paper fully disclose all the information needed to reproduce the main experimental results of the paper to the extent that it affects the main claims and/or conclusions of the paper (regardless of whether the code and data are provided or not)?
    \item[] Answer: \answerYes{} 
    \item[] Justification: We disclose all the details in Appendix~\ref{appendix:exp_details}.

\item {\bf Open access to data and code}
    \item[] Question: Does the paper provide open access to the data and code, with sufficient instructions to faithfully reproduce the main experimental results, as described in supplemental material?
    \item[] Answer: \answerYes{} 
    \item[] Justification: We will open-source the data and code when the paper is accepted.

\item {\bf Experimental setting/details}
    \item[] Question: Does the paper specify all the training and test details (e.g., data splits, hyperparameters, how they were chosen, type of optimizer, etc.) necessary to understand the results?
    \item[] Answer: \answerYes{} 
    \item[] Justification: We disclose all the details in Appendix~\ref{appendix:exp_details}.

\item {\bf Experiment statistical significance}
    \item[] Question: Does the paper report error bars suitably and correctly defined or other appropriate information about the statistical significance of the experiments?
    \item[] Answer: \answerYes{} 
    \item[] Justification: We show the results' mean and standard deviation.

\item {\bf Experiments compute resources}
    \item[] Question: For each experiment, does the paper provide sufficient information on the computer resources (type of compute workers, memory, time of execution) needed to reproduce the experiments?
    \item[] Answer: \answerYes{} 
    \item[] Justification: We disclose all the details in Appendix~\ref{appendix:exp_details}.
    
\item {\bf Code of ethics}
    \item[] Question: Does the research conducted in the paper conform, in every respect, with the NeurIPS Code of Ethics \url{https://neurips.cc/public/EthicsGuidelines}?
    \item[] Answer: \answerYes{} 
    \item[] Justification: We have adhered to the NeurIPS Code of Ethics in the paper.

\item {\bf Broader impacts}
    \item[] Question: Does the paper discuss both potential positive societal impacts and negative societal impacts of the work performed?
    \item[] Answer: \answerYes{} 
    \item[] Justification: We have discussed the broader impacts of our work in Appendix~\ref{appendix:broader}.
    
\item {\bf Safeguards}
    \item[] Question: Does the paper describe safeguards that have been put in place for responsible release of data or models that have a high risk for misuse (e.g., pretrained language models, image generators, or scraped datasets)?
    \item[] Answer: \answerNA{} 
    \item[] Justification: The paper poses no such risks.

\item {\bf Licenses for existing assets}
    \item[] Question: Are the creators or original owners of assets (e.g., code, data, models), used in the paper, properly credited and are the license and terms of use explicitly mentioned and properly respected?
    \item[] Answer: \answerNA{} 
    \item[] Justification: The paper does not use existing assets.

\item {\bf New assets}
    \item[] Question: Are new assets introduced in the paper well documented and is the documentation provided alongside the assets?
    \item[] Answer: \answerYes{} 
    \item[] Justification: We will open-source the new assets for training.

\item {\bf Crowdsourcing and research with human subjects}
    \item[] Question: For crowdsourcing experiments and research with human subjects, does the paper include the full text of instructions given to participants and screenshots, if applicable, as well as details about compensation (if any)? 
    \item[] Answer: \answerNA{} 
    \item[] Justification: The paper does not involve crowdsourcing nor research with human subjects.

\item {\bf Institutional review board (IRB) approvals or equivalent for research with human subjects}
    \item[] Question: Does the paper describe potential risks incurred by study participants, whether such risks were disclosed to the subjects, and whether Institutional Review Board (IRB) approvals (or an equivalent approval/review based on the requirements of your country or institution) were obtained?
    \item[] Answer: \answerNA{} 
    \item[] Justification: The paper does not involve crowdsourcing nor research with human subjects.

\item {\bf Declaration of LLM usage}
    \item[] Question: Does the paper describe the usage of LLMs if it is an important, original, or non-standard component of the core methods in this research? Note that if the LLM is used only for writing, editing, or formatting purposes and does not impact the core methodology, scientific rigorousness, or originality of the research, declaration is not required.
    \item[] Answer: \answerNA{} 
    \item[] Justification: The core method development in this research does not involve LLMs as any important, original, or non-standard components.

\end{enumerate}

\end{document}